\documentclass[conference]{IEEEtran}
\IEEEoverridecommandlockouts

\usepackage[utf8]{inputenc}

\usepackage{graphicx}
\usepackage{adjustbox}
\usepackage{amsmath}
\usepackage{centernot}
\usepackage{multirow}
\usepackage{booktabs}
\usepackage{subfigure}
\usepackage{caption}
\usepackage{dsfont}
\usepackage{enumerate}
\usepackage{enumitem}
\usepackage{empheq}
\usepackage{amssymb}
\usepackage{amsfonts}
\usepackage{totcount}
\usepackage{makecell}
\usepackage[table]{xcolor}
\usepackage{changepage}
\usepackage{stfloats}
\usepackage{cite}
\usepackage{soul}
\usepackage[ruled,vlined]{algorithm2e}
\usepackage{algpseudocode}

\ifCLASSOPTIONcompsoc

\usepackage[nocompress]{cite}
\else
  \usepackage{cite}
\fi

\ifCLASSINFOpdf

\else

\fi

\begin{document}
%
\title{Incorporating Causal Effects into Deep Learning Predictions on EHR Data
\thanks{Supported by NSF \#1602394, \#1934600, and NIH/NIGMS 5R01GM120079.}
\vspace{-2.5mm}
}

\newcommand*\samethanks[1][\value{footnote}]{\footnotemark[#1]}
\author{Jia Li${}^*$\thanks{*University of Minnesota, Department of Computer Science and Engineering. \{jiaxx213,yang6993,kumar001,stei0062\} @umn.edu; $^\dagger$University of Pittsburgh, Department of Computer Science. xiaowei@pitt.edu; $^{\ddagger}$University of Minnesota, Institute for Health Informatics. simo0342@umn.edu}, Haoyu Yang$^*$, 
Xiaowei Jia$^{\dagger}$, 
Vipin Kumar$^*$, 
Michael Steinbach$^*$, Gyorgy Simon$^{\ddagger}$ 
}

\maketitle

\begin{abstract}
Electronic Health Records (EHR) data analysis plays a crucial role in healthcare system quality. Because of its highly complex underlying causality and limited observable nature, causal inference on EHR is quite challenging. Deep Learning (DL) achieved great success among the advanced machine learning methodologies. Nevertheless, it is still obstructed by the inappropriately assumed causal conditions. 
This work proposed a novel method to quantify clinically well-defined causal effects as a generalized estimation vector that is simply utilizable for causal models. We incorporated it into DL models to achieve better predictive performance and result interpretation.
Furthermore, we also proved the existence of causal information blink spots that regular DL models cannot reach.
\end{abstract}


%
\IEEEpeerreviewmaketitle

\section{Introduction}
Given the wide adoption of Electronic Health Records (EHR) systems in the US, there is a surge of interest in building EHR data-driven models for predicting clinical events, and it is still ongoing. 
They are used to estimate disease progress or identify the risky population, enabling better diagnosis, prognosis, and treatment decisions. Hence their predictive accuracy, as well as the result interpretations, are both crucial for healthcare quality,
especially in the context of real-time applications at the bedside, like precision medicine \cite{deigner2018precision}.

It is known that clinical processes are motivated by the underlying causal mechanism and lead to observable events along the timeline, e.g., with some medication through organs,
the effects sequentially come to observation.
But events will cause subsequent events and produce much-complicated relations, such as mutually determined disease progress and medical upgrading.
Therefore, people have widely adopted causality modeling techniques in EHR analysis in the past decade, which has been well developed as an essential branch of Machine Learning (ML) \cite{wu2010prediction}.

However, challenges still exist.
In practice, lots of clinical events can only be limited to observable. Thus, information in EHR may not be sufficient to reflect the underlying causation fully.
The significant application that aims to discover causal information as much as possible from observations is known as causal inference.
This area has attracted various applications of advanced learning techniques, such as DL methods \cite{miotto2018deep, shickel2017deep}. 
Based on Neural Networks (NNs) architecture, DL models have shown great success in harnessing the power of extensive EHR data and coming up with better performance. This success is thanks to their advanced ability to extract deep hierarchical features and non-linear dependencies.

Nevertheless, modeling graphical causality relies on certain causal assumptions, and so do the DL-based models. They may lead to inevitable deviations in causal model results and their interpretations.

On the other hand, health statistics have provided extensive knowledge about the hidden causation, but they are not intuitively utilizable for DL models. We notice that among the applications on EHR, DL models are bare to be considered as short of the capability to discover causal effects. In other words, questions are still unclear: In which scope the causal knowledge is learnable by integral modeling? And does there exist DL-irreplaceable knowledge in domain statistics?

To answer these questions, we propose a novel method to incorporate clinically well-defined causal effects into DL models and check their predictive performance improvement. We hypothesize that some causal information exists that regular DL models cannot automatically discover. And they need external knowledge infused to reinforce their causal learning ability.

We summarize our contributions as:
\begin{enumerate}
    \item Proved existence of the causal-info blink spots for DL.
    \item Raised a novel method to quantify the domain-defined causal effects as a generalized estimation vector.
    \item Realized a generally usable and configurable Causal EHR data Generator and successfully used it to simulate a complex comorbidities progress trajectory.
\end{enumerate}

\section{Related Work}\label{sec:background}
EHR data attributes usually include diagnosis, lab results, medical notes, prescriptions, along with demographics (age, gender, race) and other clinically relevant features.
The various data types make it richly informative and challenging to build up the causal model.

In this section, we will introduce structural causal models with DL-based techniques and their limitations on EHR data; and then briefly summarize the current DL applications in EHR studies.

\vspace{-2mm}
\subsection{Graphical Causal Models with DL}
\label{sec:rw_cd}
In the last decade, the interest in combining DL models with causality learning has increased in both the DL and the causality communities \cite{young2020learning}.
These tasks can be briefly classified as causal structure learning, causal effects estimation, or on a mixed purpose of both.

The methods that learn causality as an integral structural model are known as "Structural Equation models" (SEMs), or "Functional Causal Models" (FCMs) \cite{
goudet2018learning, lawrence2021data}. 
The structure is commonly described as conditionally-dependent relationships among variables using Directed Acyclic Graphs (DAGs).
Usually, the parametric methods assume deterministic functions for all direct causal relations and apply constraints globally on the joint distribution \cite{glymour2019review}.

On the other hand, the advantage of NNs-based methods is on generalized non-parametric modeling and being able to avoid prior assumptions.
But due to the short of structural consideration, they are commonly applied on causal effects estimations within a determined or non-required causal structure \cite{louizos2017causal, bica2020estimating, chattopadhyay2019neural}.
In recent years, a methodology arose to transform the discrete graphical structure learning problems into continuous optimization ones that make NNs applicable \cite{luo2020causal}.
The key idea is to quantify the acyclic-ness that is optimizable by a constrained adjacency matrix.

However, the learned causal relations do not necessarily have a causal interpretation.
Mostly, they rely on or imply the existence of \emph{Causal Sufficiency} Assumption or \emph{Causal Markov} Assumption that are possibly unverifiable \cite{judea2010introduction}.
\vspace{1.5mm}\\
\emph{(1) Causal Sufficiency.}

It assumes that the DAG of modeling has included all causal relations among variables. In some scenarios, it can also be referred to as Causality Faithfulness.

Indeed, the conditional dependency of DAG can only provide necessary but not sufficient conditions for identifying causalities.
There may exist other relations of interest, not entailed by the structural modeling but assumed to be nonexistent. 
\vspace{1.5mm}\\
\emph{(2) Causal Markov Condition}.

It means that every node in the DAG is conditionally independent of its nondescendents, given its parents, which has been
generally acknowledged short of the capability to distinguish confounding.

\vspace{1.5mm}
In practical applications, deep learning models are much more potent than the conventional methods of discovering hidden causal relations. But since they still have applied inappropriate assumptions and black-box nature, they can hardly guarantee to produce interpretable results.

\begin{figure}[h!]
\vspace{-4mm}
\centering
\includegraphics[scale=0.45]{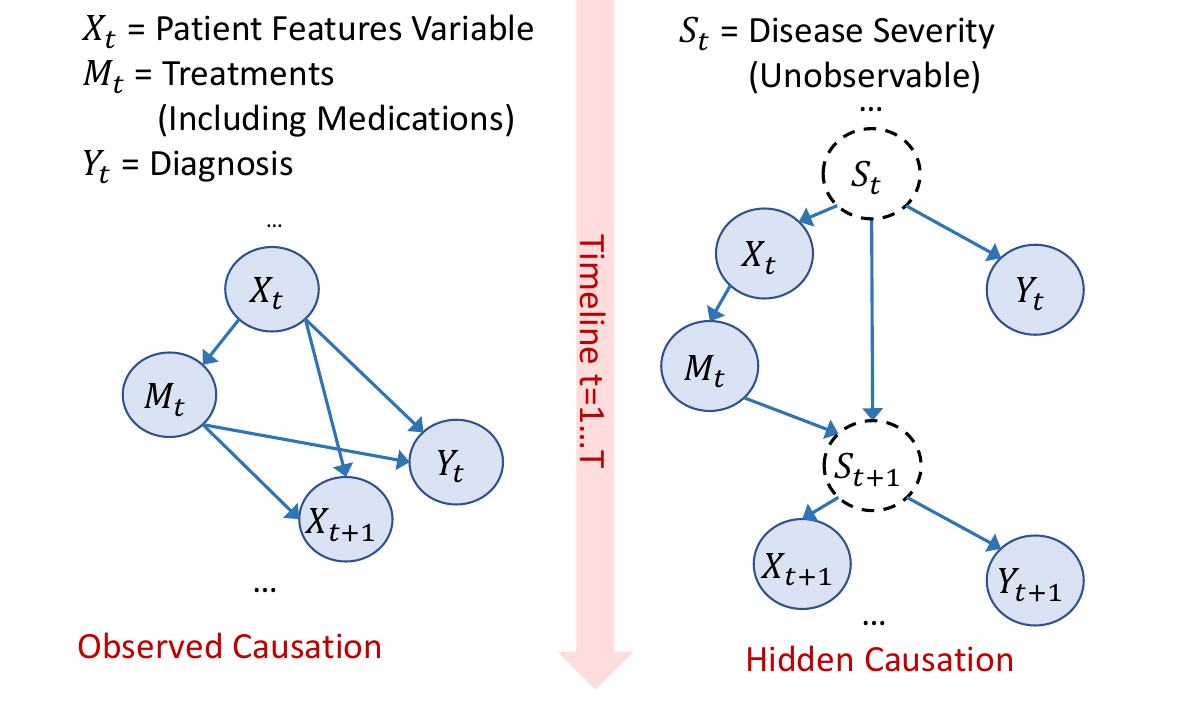}
\vspace{-6mm}
\caption{The EHR Causal Structure. Left: observed view. Right: knowledge-based inferred view. The solid circle means element observable, and the dashed one tells not.}
\label{fig:underlying_causal}
\vspace{-3mm}
\end{figure}

\vspace{-3mm}
\subsection{DL Applications on EHR}
In the recent five years, DL-based methods turned out to be favorable for EHR data applications.
They are primarily used in two categories: prediction tasks and representation learning. Here we mainly concentrate on the predictions.

DL models are commonly adopted to enhance predictive performance effectively.
For example, Deep Patient \cite{miotto2016deep} predicts multi-outcomes of patients, Deeper \cite{wickramasinghe2017deepr} targets on hospital re-admissions, and Doctor AI \cite{choi2016doctor} was proposed for heart failure prediction. 

For temporal EHR data, RNN is mainly considered as an advanced time-series model \cite{esteban2016predicting, che2018recurrent, aczon2017dynamic}. There are two popular mechanisms of RNNs, Gated Recurrent Units (GRU) \cite{choi2016doctor, choi2016retain} and Long Short-term Memory (LSTM), \cite{rajkomar2018scalable, suresh2017clinical} that employ different network architectures but are similarly efficient to leveraging temporal information.

However, EHR data has particular characteristics that may mislead DL models.
Considering their applications on treatment effects learning, we raise a hypothesis that RNNs cannot automatically discover the underlying causal information of interest and may need external knowledge infusion in some ways.

\vspace{-2mm}
\section{Causality on EHR Data}
\label{sec:ehr_cause}
This section mainly discusses the mechanism of underlying causality in EHR. And then, from the aspects of causal assumptions, we use living examples to illustrate the reason for their violations.

\subsection{Underlying Causal Structure}
Suppose an EHR dataset spans $T$ time steps.
At any step $t$, the observed data elements include: recorded patient attributes $X$ (e.g., lab results, vital signs, and demographics), subscribed medications $M$ (or some other clinical treatments), and disease diagnosis $Y$.

Fig.~\ref{fig:underlying_causal} displays the typical causal structure from two views: the one directly observable from data and the perspective based on background knowledge.

Here the hidden variable $S_t$ represents patient's \emph{Disease Severity} at time $t$.
Without loss of generalization, we define it as aggregating all helpful information to infer the causal relations at $t$. As $X_t$ only indicates the visible part, we generally consider $S_t$ more informative than $X_t$.

Commonly, the medical plans of patients are decided by physicians' observation, including the historical records denoted as $X_t$ here. Then the medication effects reflect on the subsequent records $X_{t+1}$, which form a typical confounding relationship. The scientists from traditional statistics have adequately explored solutions to perform de-confounding that involve extensive clinical knowledge.

However, a more significant challenge for predictions on EHR is the unobservable critical information. The disease severity does not have a specific clinical definition but only acts as a conceptual element in our models. Among the machine learning literature, we have ever reviewed, mostly $X_t$ is simply considered the roll of $S_t$ without other choices. But that essentially leads to some problems, like the violation of regularly applied causal assumptions. 

On the other hand, the experts from health stats have revealed that although the disease severity is not directly observable, they can be inferred, or at least partially inferred from the observations by leveraging domain knowledge.
Therefore, the motivation of this work is to combine the advantages of generalizable DL methods and interpretable statistical models.

\vspace{-1mm}
\subsection{Invalid Causal Assumptions on EHR}
\vspace{-1mm}
Because of the omitted underlying information, the causality learned of EHR may be distorted to some extent and possibly leads to violations of standard causal assumptions.

\vspace{1mm}
\noindent
\emph{(1) Unreliable Criterion - Causal Sufficiency violated.} \\
Usually, disease criterion is set up as a metric of certain test results. But it may be triggered by other causes and lead to false positives.

\noindent
\texttt{\small{\textbf{[Example A]}. For diabetes patients, high Blood Pressure (BP) is a sign of danger. But in cardiovascular surgery (which may be caused by severe diabetes), increasing BP by anticoagulant drugs is mandatory to prevent blood clotting. 
}}

\vspace{1mm}
\noindent
\emph{(2) Mislead Medicine - Causal Sufficiency violated.} \\
Adding more intense medication is usually considered worse ill, but the reality may be counter-intuitive. 

\noindent
\texttt{\small{\textbf{[Example B]}.
Increasing insulin use is a significant sign of worse diabetes. But after a kidney transplant surgery (may be caused by severe diabetes), this increasing conversely indicates a more functional body.}}

\vspace{1mm}
\noindent
\emph{(3) Non-negligible History - Causal Markov Condition violated.} \\
Similar records may be developed as different disease progresses.

\noindent
\texttt{\small{\textbf{[Example C]}.
Pregnant women who just developed gestational Hypertension (HTN) has very similar lab results with an HTN patient who just got pregnant but differentiate on personal effects to anti-hypertension drugs.
}}

\section{Proposed Methodology}
\label{sec:meth}
To bring usable knowledge into the DL modeling process, we propose a causal effect estimator, calculated with the traditional method but produce new information that can hardly be discovered by regular DL models, and eventually increase the predictive performance.
We must emphasize that our calculation does not rely on any extra data input.

We will first introduce the definition, present the calculation procedure, and finally, analyze the fundamental principle.

\subsection{Treatment Effect Estimator $\Delta$}\label{subsec:Ddef}
We use $X_t$ to denote the patients observation records at time $t$ that are predictive to the next time step outcome $Y_{t+1}$ or records $X_{t+1}$.
Practically, $X_t$ consists of lab results and vital signs, such as blood pressure and BMI index. 
Then, we use $M_t$ to denote the treatments of interest at time $t$. In our context, $M_t$ is the vector of all possible medications, but it can also include surgeries, therapies, or any other clinical treatments in practice.

Let vector $M_t=(m_1,m_2,\ldots,m_n)$, where $ m_i \in [0,1]$. $m_i$ indicates whether the $i$th medication (out of $n$ in total) is applied to patient at time $t$ ($m_i=1$) or not ($m_i=0$).
We use $M_t^{i+}$ and $M_t^{i-}$ to represent pair of the two possibilities at time $t$ that $M_t^{i+}$ has $m_i=1$ while $M_t^{i-}$ has $m_i=0$, and all the other medications keep identical between them.
To be specific, $M_t^{i+} = (m_1,m_2,\ldots,m_i=1\ldots,m_n)$ and $M_t^{i-} = (m_1,m_2,\ldots,m_i=0\ldots,m_n)$.

The considered scenario is that given known $M^{i-}_{t-1}$ at time $t-1$, the patient will possibly be under ongoing treatment as $M^{i-}_{t}$ or additionally medicated as $M^{i+}_{t}$.
Accordingly, we use $X^{i+}_{t+1} = (X_{t+1}|M_t^{i+})$ and $X^{i-}_{t+1} = (X_{t+1}|M_t^{i-})$ to represent two possible values of the observable records at $t+1$, under the two different medical plans at $t$, respectively.

\vspace{1mm}
\emph{\textbf{Definition 1}. The \underline{Treatment Effect Estimator} $\Delta^i_{t+1}$ about time $t$, is defined as \\
}
$\Delta^i_{t+1}=
\begin{cases}
  (X^{i+}_{t+1}-X^{i-}_{t+1}), & \text{if}\ \mathds{1}(M_t^{i+})=1 \text{, given } \mathds{1}(M_{t-1}^{i-})=1 \\
  0, & \text{otherwise}
\end{cases}$ \\

Here we only discussed the situation with a single medication $m_i$ possibly changed. However, it can stand for any medications subset of interest, as the user-determined.

\smallskip
The proposed estimator $\Delta^i_{t+1}$ can be described as ``how much $X$ would be changed if the additional medications were applied, compared with not applied.''
Apparently, $\Delta^i_{t+1}$ can only have non-zero meaningful value when $M_t^{i+}$ is known to be happened. In other words, at time $t$ the truly observed \emph{Factual Value} is $X^{i+}_{t+1}$, while $X^{i-}_{t+1}$ is called as \emph{Counterfactual Value}.

The idea about $\Delta$ is inspired by a well-grounded conception from clinical studies, the \emph{Individual Treatment Effect} (ITE) \cite{shalit2017estimating}, which is defined as the individual-level difference between the factual and counterfactual values of the outcome for each patient, such as some diagnosis or mortality risk. Instead of outcomes, we creatively use it upon data attributes to form a ``difference vector'' to describe the medical effect. 


\subsection{Estimation of $\Delta$ Value}\label{subsec:Dval}
By $\Delta$'s definition, the challenge to calculate it is estimating the counterfactual values, which are not explicitly observable from data.
Since $\Delta$ can have a non-zero value only when $M_t^{i+}$ is true, the estimation task can be defined as: finding the counterfactual value for each patient who certainly has $M_{t-1}^{i-}$ at $t-1$ and $M_t^{i+}$ at $t$.

Intuitively, the so-called counterfactual value is the outcome value of a ``counterfactual case'' if it exists. Thus we can search through the entire dataset and find a record ``matched'' with the target record but turns out to choose the alternative medical plan at time $t$, and then we take its value at time $t+1$ to be the desired counterfactual $X^{i-}_{t+1}$. Ideally, the matched records pair should have identical values at $t-1$, which are the conditions for making medical decisions at $t$. However, in practice, we need to figure out how to find a record as matched as possible.

In clinical studies, a conventional method, \emph{Propensity Score Matching} (PSM), is widely used to isolate the unbiased treatment effects. Here ``isolate'' means selecting a subpopulation from the cohort, in which the different medical plans are decided in a uniform probability. In other words, the ``randomized clinical trial'' is strategically picked out from the actual patients' data.
To be specific, the selection strategy is to find pairs of matching records, both of which have the same chance to be assigned $M_t^{i+}$ given their similar $X^{i-}_{t-1}$, but only one turns out to be $M_t^{i+}$ while the other gets $M_t^{i-}$ at time $t$.

\vspace{1mm}
\emph{\textbf{Definition 2}. 
Given record $A$ with $M_{t}^{A} \neq M_{t-1}^{A}$, record $B$ is \underline{Counterfactual Record} of $A$, \ 
if and only if \ $M_{t}^{B} = M_{t-1}^{B} = M_{t-1}^{A}$\ 
and \ $\mathbb{P}(M_t^{A}|X^A_{t-1}) \approx \mathbb{P}(M_t^{A}|X^B_{t-1})$.}  
\vspace{1mm}

To simplify the notation, here we use $M_{t}^{A}$ and $X_t^{A}$ to denote medications vector and the record value of $A$ at time $t$, and as the same way, $M_{t}^{B}$ and $X_t^{B}$ are $B$ notations. $\mathbb{P}$ stands for probability; and the approximate equation means that $A$ and $B$ have their $X_{t-1}$ values to be approximately same predictive about $M^A_{t}$.

In this work, the conditional probability about $M^A_t$ is estimated as a predictive logistic regression. The desired $\Delta$ is calculated as $\Delta_{t+1} = X^A_{t+1} - X^B_{t+1}$. In practice, it is unnecessary to get an estimation for every possible $\Delta_{t+1}$, which could be exponentially numerous. A meaningful $\Delta$ is as to represent a treatment effect of interest. For the example in Definition 1, $\Delta^i_{t+1}$ is estimated as the conditional effect of medical plan $M_t^{i+}$, given the ongoing medical plan $M_{t-1}^{i-}$.

\subsection{Functional Theory Analysis}\label{subsec:Drole}
This section will theoretically analyze the functional principle of the proposed method. At first, $\Delta$ contains information that can hardly be discovered in sequentially learning. 
And then, $\Delta$ has the nature to provide attention guidance to NNs, and also be intuitively configurable.

\subsubsection{$\Delta$ reveals new information}{.\\}

\begin{figure}[h!]
\vspace{-3mm}
\centering
\includegraphics[scale=0.37]{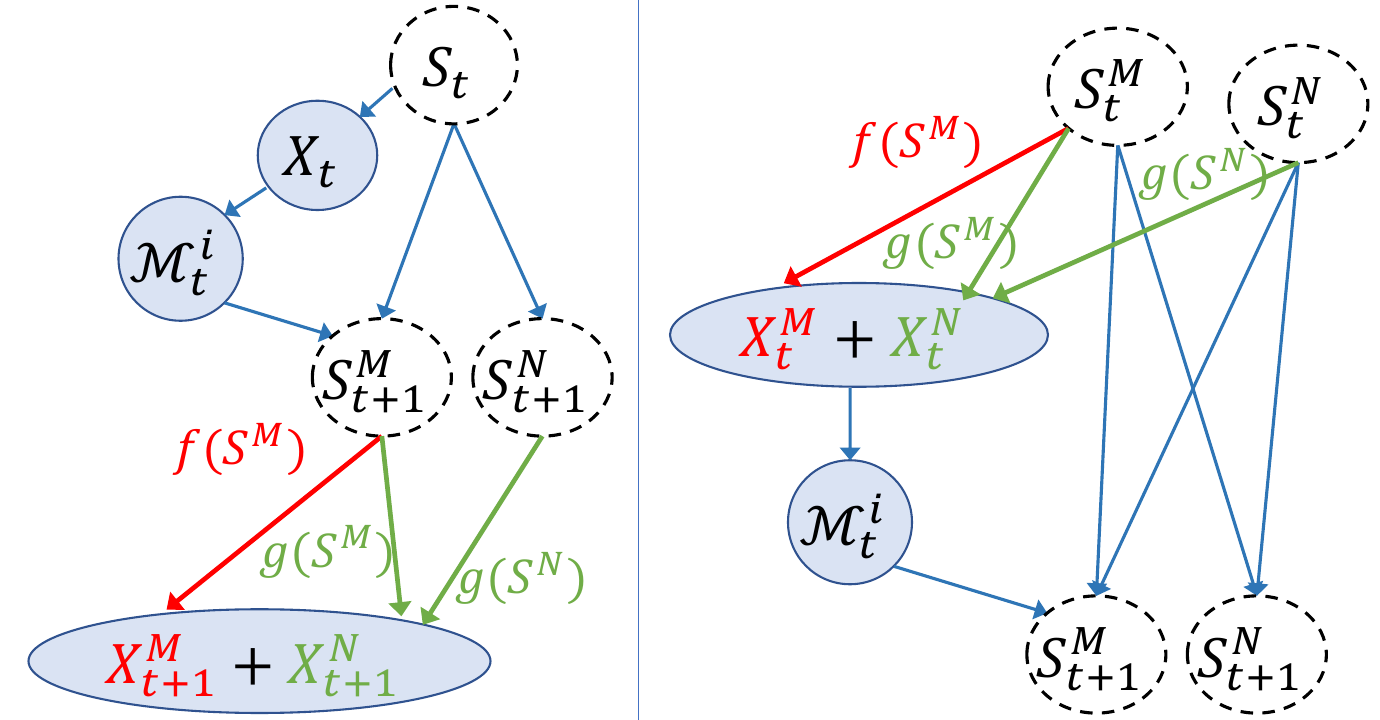}
\caption{Disease Progress Causal Graph. The solid outline means element observable, and the dashed one means not. Here the node $M^{i+}_t$ stands for its binary indicator $\mathds{1}(M^{i+}_t)$.
}
\label{fig:add_Delta}
\end{figure}

\vspace{-3mm}
We have assumed the not-observable variable of disease severity, denoted as $S_t$, to fully include information that could be deterministic to any observed variable at time $t$.
The left side of Fig.\ref{fig:add_Delta} shows the disease progress over time with unobserved $S_t$ and $S_{t+1}$ involved.
The observed data elements $X_t$ and $X_{t+1}$ represent the patients observation records at time $t$ and $t+1$ respectively; and $M_t$ is the medications vector, also referred as clinical treatments; The outcomes $Y$ of interest are downstream from $S_{t+1}$.





In the full model, the disease severity $S_t$ is decomposed into two components $S_t^N$ that models the natural progression and $S_t^M$ that adjusts for medications. 
The right side of Fig.\ref{fig:add_Delta} shows the  correspondingly changed causation.
This decomposition aims to isolate the severity adjustment $S_t^M$ that medications $M_{t-1}$
can only decided, and leaves $S_t^N$ independent with $M_{t-1}$, which is the definition of pure effect from treatment $M_{t-1}$ upon the unobservable severity $S_t$.
In Fig.\ref{fig:add_Delta}, the element $S_{t+1}^N$ is independent with $M_t^{i+}$ for any time $t$ .

\vspace{-2.5mm}

The observations $X_t$ at the same time $t$ are also composed of two parts $X_t^M$, and $X_t^N$, separated on the similar principle that $X_t^N$ stands for the observed values naturally progressed without medication, and $X_t^M$ adjusts the values for medicine applied.

Notably, the severity component $S_t^N$ and $S_t^M$ cannot be assertively assumed independent of the observed values on the other side, i.e., $X_t^M$ and $X_t^N$, respectively. 
The decomposed causation has four links at most, as presented in Fig.\ref{fig:add_Delta}.
For a straightforward view, we differentially color the $S_t^M$-started and the $S_t^N$-started ones as green and blue respectively. Moreover, we name the causal functions toward $X_t^M$ as $f$, while $X_t^N$ as $g$, and each one comprises two individual terms that $f=(f_1,f_2)$ and $g=(g_1,g_2)$.
$f_1,f_2,g_1,g_2$ can be arbitrary functions, thus they represent the generalized model that does not require any prior assumption about distribution. However, not all of the four are necessary for the model.

\newcommand{\indep}{\perp \!\!\! \perp}

\smallskip
\emph{\textbf{Theorem 1}. Function $f_2$ is constant, and
$X^M_{t} \indep S^N_{t}$.}

\smallskip
\emph{{Proof}:}
It has been known that $X^M_{t}=f(S_t)=f_1(S^M_{t}) + f_2(S^N_{t})$. And then, by definition we knows $X^M_{t}=0$ iff $S^M_{t}=0$. Therefore, we have $f_2(S^N_{t}) = - f_1(0)$ to be constant, i.e.  $X^M_{t} \indep  S^N_{t}$.
\smallskip





According to Theorem 1, function $f_2$ is dashed in Fig. \ref{fig:add_Delta} to indicate its unnecessity. 
We attempt to estimate observations at time $t+1$, including $X_{t+1}$ and $Y$, by only given the observable elements at time $t$; at the same time, $M^{i+}_t$ is supposed to be happened, i.e. $\mathds{1}(M^{i+}_t)=1$.

To simplify the estimation of $X_{t+1}$, let us firstly convert $X^N_{t+1}$ as an expression of $X^M_{t+1}$, and then concentrate to estimate $X^M_{t+1}$.

By definitions and Theorem 1, we know that $X^M_{t} = f_1(S^M_{t})$, and $X^N_{t} = g_1(S^M_{t}) + g_2(S^N_{t}) = g(S_t)$.
Thus we have the expression for $X_t^N$ and also its isomorphic expression about $X_{t+1}^N$:
\vspace{-.5mm}
\begin{align}\label{eq:fun1}
\begin{split}
X^N_{t} & =  g\ ( f_1^{-1}(X^M_{t}),\  S^N_{t})
\end{split}
\end{align}

\vspace{-5mm}
\begin{align}\label{eq:fun2}
\begin{split}
X^N_{t+1} & =  g\ ( f_1^{-1}(X^M_{t+1}),\  S^N_{t+1})
\end{split}
\end{align}
\vspace{-.5mm}
Given $\mathds{1}(M^{i+}_t)=1$, the non-zero valued $S^M_{t+1}$ has decomposition as:
\vspace{-1mm}
\begin{equation}\label{eq:fun3}
S^M_{t+1}  = \alpha S^M_t + \beta S^N_t
\end{equation}
\vspace{-1mm}
where $\alpha$ and $\beta$ are the linear transformation matrices for describing two components of the severity temporal progression: one is from $S^M_t$ to $S^M_{t+1}$ and the other is from $S^N_t$ to $S^M_{t+1}$. Then we have:
\vspace{-1mm}
\begin{align*}\label{eq:fun4}
\begin{split}
X^M_{t+1} =  f_1(\alpha S^M_t + \beta S^N_t ) = \alpha X^M_t + f_1(\beta S^N_t) \\
\end{split}
\end{align*}
\vspace{-1.5mm}
that can be converted as the expression of $S^N_t$:
\vspace{-1mm}
\begin{equation}\label{eq:fun4}
S^N_t = \frac{1}{\beta} f_1^{-1}(X^M_{t+1} - \alpha X^M_t)
\end{equation}
\vspace{-1.8mm}
By substituting Eq.(\ref{eq:fun4}) into Eq.(\ref{eq:fun1}) we have:
\vspace{.5mm}
\begin{equation}\label{eq:fun5}
X^M_{t+1} = (\alpha -\beta)X^M_t + \beta f_1(g^{-1}(X^N_t))
\end{equation}

\vspace{-1mm}
From Eq.(\ref{eq:fun5}) we know that $X^M_{t+1}$ is estimable given only the observable variables $X^M_{t}$ and $X^N_{t}$ without requiring the hidden values of severity.
On the other hand, from Eq.(\ref{eq:fun2}) we find that the estimation of $X^N_{t}$ still needs value of $S^N_{t+1}$, whose isomorphic expression about $S^N_{t}$ is shown as Eq.(\ref{eq:fun4}). From Eq.(\ref{eq:fun4}) we can deduce that $S^N_{t+1}$ will become estimable when the following $X^M_{t+2}$ is observed, but not yet at the current time $t+1$.

In short, $X^M_{t+1}$ can be estimable if $X^M_{t}$ and $X^N_{t}$ are known, while $X^N_{t+1}$ is not directly estimable thus can only be deduced by $X^N_{t+1} = (X_{t+1} - X^M_{t+1})$ with known $X_{t+1}$ and an estimated $X^M_{t+1}$ at time $t+2$. 

In practical prediction tasks, values of $X_{t}$, $X_{t+1}$, $X_{t+2}$, $...$ will successively become observed along the timeline. Therefore, for any time $t$ the value of $X^N_{t+1}$ will be eventually estimable, if and only if the splitting between $X^M_{t}$ and $X^N_{t}$ is known for each $X_{t}$.

\smallskip
\emph{\textbf{Theorem 2}. $\Delta_t$ reveals necessary information for estimating the observable treatment effect $X^M_{t+1}$.}

\emph{{Proof}:}
By Definition 1, we have
\vspace{-1.5mm}
\begin{align*}
\begin{split}
\Delta_{t+1} & =  (X_{t+1}|M_t^{i+})-(X_{t+1}|M_t^{i-})\\
& = [X^M_{t+1} + X^N_{t+1}] - x^N_{t+1} 
= X^M_{t+1}
\end{split}
\vspace{-1mm}
\end{align*}
and subsequently have $\Delta_{t} = X^M_{t}$ due to the isomorphism.

This equation indicates $\Delta_{t}$ to be an effective estimation of $X^M_{t}$, and $X^N_{t}$ is equivalently estimated. Thus, $\Delta_{t}$ reveals the critical information for estimating the follow-up $X^M_{t+1}$ and $X^N_{t+1}$.

\vspace{1.5mm}
With the value of $\Delta_{t}$ estimated by our method, the performance of predicting outcome $Y$ depends on how capable the model is to learn the general causal functions $f_1, g, \alpha, \beta$, and also the ones started from severity $S$ toward $Y$.
Notably, in both theorems we use $g$ as an integral function instead of comprising two separable components like $g=(g_1,g_2)$. That implies $(S^M, S^N)$ to be possibly associated and jointly predictive of $X^N$, without assuming independently with each other.

Compared to parametric regressions, DL methods are more potentially powerful to be improved performances since, without causality distribution assumptions, they are much better at modeling generalization.


\newcommand{\bigCI}{\mathrel{\text{\scalebox{1.07}{$\perp\mkern-10mu\perp$}}}}
\newcommand{\nbigCI}{\centernot{\bigCI}}

\subsubsection{$\Delta$ can guide NNs attention}{.\\}

\vspace{-2mm}
Inherently, the proposed method faces the challenge of interesting causal effects selection.
Per definition, $\Delta$ is estimable between a pair of medical plans that have an arbitrary subset of the medications to be different.
Thus the theoretically possible $\Delta$ could be exponentially numerous if the considered medications vector is within a comparatively large length or dimension.

But in practical problems, causal effects of interest are more likely to be predefined and in a reasonable total number for intuitively understandable. 
In EHR data analysis, the primarily selected modeling objective is one specific or a few relative diseases, like a group of comorbidities; and the treatment effects of interest are often about adding the medications of interest, given the ongoing medical plan.
So the number of meaningful $\Delta$ is practically limited.

Consequently, the estimated $\Delta$s will highly possible to be a sparse matrix, while the empty elements are left as zeros.
The non-zero elements with larger absolute values indicate their corresponding treatment effects to be more comparatively significant, including both positively and negatively.

In the input layer of NNs, the sparse $\Delta$s attributes can very suitably work as the attention vector to guide the optimization processing toward more efficient direction and zero out the misleading or ignorable factors.

Furthermore, in case of inappropriate causal assumptions, $\Delta$ can be manually customized to avoid misled modeling. As in the examples mentioned above:

\texttt{\small{\textbf{[Example A]}. At surgery time, zero out the BP relative digits in $\Delta$ to avoid wrong attention.}}

\texttt{\small{\textbf{[Example B]}. the doctor can set the effects of adding insulin dose as positive or negative in $\Delta$}}.

\texttt{\small{\textbf{[Example C]}. $\Delta_{t}$ can bring historical information since it is reflecting the changes that happened at $t-1$ }}.

\section{Experimental Design}
\label{sec:design}
The experiments aim to confirm $\Delta$ to be significantly informative by comparing the predictive performances with and without $\Delta$.

For convincing conclusions, we employed two type datasets to perform experiments: (1) a real-world EHR dataset of type-II diabetes patients; (2) multiple synthetic datasets generated with underlying causation, determined by random DAGs. 

Due to the practical limitation of actual data (e.g., missing values), the advantage of $\Delta$ cannot be fully revealed on it. That's why synthetic data is necessary.

This section will introduce the two types of datasets and the experimental tasks on them, respectively. The synthetic data generation method will be illustrated, and we have published the source code for free downloading. 

\subsection{Real EHR Data}
Mayo Clinic (MC) provides primary care to residents of Olmsted County, Minnesota, and has an integrated electronic health record system including diagnoses, medications, laboratory results, and clinical notes.

With research consent, we used a retrospective cohort of de-identified data from 73,045 primary care patients at Mayo Clinic, Rochester, MN. The cohort consists of patients aged $\ge$18 and $\le$ 89 at baseline on Jan. 1st, 2005, having at least one visit before and after baseline. These patients were followed until 2017 (median follow-up time is ten years). We extracted patient demographics and diagnoses (ICD-9), laboratory results, vital signs, and medications longitudinally for six non-overlapping time windows: before-2004, 2005-2006, 2007-2008, 2009-2010, 2011-2012, 2013-after. For the outcomes that happened in time window t, we make predictions based on observations from previous time windows, i.e.,  t-1, t-2, etc. The latest observed values are taken for the predictors in each time window.

The patient cohort description is given in Table\ref{tab:cohort}.

\begin{table}
\caption{ Study Cohort Description}
\label{tab:cohort}
  \begin{center}
    \resizebox{0.9\columnwidth}{!}{%
    \begin{tabular}{c|c|c}
    \hline
      \textbf{Variable \space\space} & \textbf{Median \space\space} & \makecell{\textbf{Interquartile 
      Range}}\\
      \hline
      Age [years]  & 45 & 31,\space 59\\
      Male [\%] & 43.4 & \space \\ 
      LDL [mg/dL] & 105 & 83,\space 130\\
      TG [mg/dL] & 130 & 91,\space 187\\
      HDL [mg/dL] & 50 & 41,\space 61\\
      SBP [Hg mm] & 120 & 108,\space 132\\
      DBP [Hg mm] & 70 & 60,\space 78\\
      FPG [mg/dL] & 101 & 92,\space 117\\
      Follow-up [years] & 10 & 6,\space 12\\
      \space & \textbf{Percent} & \makecell{\textbf{Number of 
      patients}} \\
      \makecell{Antihypertensive \\ medication} & 42.4 & 30968\\
      \makecell{Antilipemic medication} & 32.8 & 23925\\
      Progressed to DM & 16.9 & 12367\\
      Progressed to CAD & 17.9 & 13065\\
      Progressed to CKD & 7.2 & 5267\\
      Progressed to Stroke & 1.1 & 779\\
      \hline
    \end{tabular} %
    }
  \end{center}
\vspace{-2mm}
\end{table}

\subsubsection{Medications}
All referred medications for T2DM and related comorbidities have been rolled up to National Drug File Reference Terminology NDF-RT pharmaceutical subclasses.

There exist three medication classes:
T2DM (DM) drugs, Hyperlipidemia (HL) drugs, and Hypertension (HTN) drugs. The table\ref{tab:med_levels} displays all three classes in increasing levels, where a higher level is commonly for higher severity. Each class is encoded as a binary vector with one digit representing one level, and we finally concatenate all three vectors to comprise the treatment indicator ${M}$.

\begin{table}[]
\caption{ The treatment levels for medication classes }
\label{tab:med_levels}
\resizebox{\columnwidth}{!}{%
\begin{tabular}{|
>{\columncolor[HTML]{EFEFEF}}c |l|l|c|}
\hline
\cellcolor[HTML]{EFEFEF} & \multicolumn{3}{c|}{Medication Classes} \\ \cline{2-4} 
\multirow{-2}{*}{\cellcolor[HTML]{EFEFEF}\begin{tabular}[c]{@{}c@{}}Treatment\\ Levels\end{tabular}} & \multicolumn{1}{c|}{DM} & \multicolumn{1}{c|}{HL} & HTN \\ \hline
Level-1 & \begin{tabular}[c]{@{}l@{}}1st non-insulin\\ drug\end{tabular} & \begin{tabular}[c]{@{}l@{}}statin as \\ the 1st drug\end{tabular} & \multicolumn{1}{l|}{1st drug} \\ \hline
Level-2 & \begin{tabular}[c]{@{}l@{}}2nd or more \\ non-insulin drug\end{tabular} & \begin{tabular}[c]{@{}l@{}}non-statin drug\\ as the 1st drug\end{tabular} & \multicolumn{1}{l|}{2nd drug} \\ \hline
Level-3 & \begin{tabular}[c]{@{}l@{}}insulin added\\ as the 1st drug\end{tabular} & \begin{tabular}[c]{@{}l@{}}non-statin drug\\ added to statin\end{tabular} & \multicolumn{1}{l|}{more drugs} \\ \hline
Level-4 & \begin{tabular}[c]{@{}l@{}}insulin added to \\ non-insulin drug(s)\end{tabular} & \begin{tabular}[c]{@{}l@{}}non-statin drug \\ added to non-\\ statin drug(s)\end{tabular} & - \\ \hline
Level-5 & \begin{tabular}[c]{@{}l@{}}non-insulin drug\\ added to insulin\end{tabular} & \multicolumn{1}{c|}{-} & - \\ \hline
\end{tabular}%
}
\end{table}

\subsubsection{Prediction Tasks}
We comprehensively set up two types of tasks. One type is binary classification, and another one is continuous outcome regression, as below:


\noindent\textbf{Task 1: Disease diagnosis predictions.}
The outcomes in this task include the disease diabetes (DM) and its three complications (CAD, CKD, Stroke). We aim to predict whether the disease developed in the subsequent time window, using predictors from the immediately preceding time windows. 

For a brief, we refer to the number of preceding time windows as \emph{Time Step} or \emph{Step}. e.g., Step=1 means the predictors from the preceding time window (2011-2012); And Step=2 means from the preceding two time windows (2009-2010 and 2011-2012).

We perform four groups of experiments, with Time Step being $1,2,3,4$ respectively.

\noindent\textbf{Task 2: Lab results forecasting.}
The outcomes in this task are continuous values, including three critical lab results in the immediately subsequent time window. Same as Task 1, four groups of experiments have Time Step = $1,2,3,4$ respectively.

\vspace{1.3mm}
\subsubsection{Compared Models}

We implemented two RNN architectures in a different way of using $\Delta$: one is simply augmenting $\Delta$ as input, and another is pre-training with $\Delta$, where the latter is expected to be more efficient to utilize $\Delta$'s information. RNN-GRU and RNN-LSTM mechanisms are used, respectively. Besides, compared with RNNs, two linear models are implemented as baseline methods.

We obtain a pair of performances for each model: with and without $\Delta$ input, whose difference reflects $\Delta$'s effect.

\noindent\textbf{Model 1: Augmentation Architecture RNN.}
For each time window $t$, we concatenate ${X_t}$ with $\Delta_t$ as new predictors, then use such augmented data as input to the RNN model.

\noindent\textbf{Model 2: Unsupervised Pre-training RNN.}
An alternative approach to use $\Delta$ to guide the learning process of the RNN model:
We pre-train the RNN such that the model predicts $\Delta_{t+1}$ from $X_{t}$ at each $t$. Since the pre-training does not require labels, i.e., unsupervised, the test data can also be used.

\noindent\textbf{Model 3: Logistic Regression Model.}
$\Delta_t$ variables are simply augmented with $X_t$ as new predictors.

\noindent\textbf{Model 4: Linear Mixed-Effects Model.}
A standard method in EHR data analysis, as patients are usually categorized as subgroups, and a mixed-effect model is convenient to capture the group-level effect. 
$\Delta$ is also augmented as new predictors.

\subsection{Synthetic EHR Data}

\vspace{-1mm}
\subsubsection{Causal data Generation Method}
The synthetic EHR data generation comprises three stages: I) Initialization, II) Disease Assignment, III) Treatment Assignment.

The detailed algorithms of the three stages are provided in Appendix A of this paper as supplementary material.

The three stages are for generating three different status of data, $X_{init}, X_{dise}$ and $X_{trea}$ sequentially, where everyone is based on the preceding one. But only the last matrix $X_{trea}$ will be exported as the synthetic EHR data, and the front two are hidden ground truth. $\Delta$ matrix generated along with $X_{trea}$. 


\noindent\textbf{Stage I: Initialization.}
Randomly set up the healthy values for each lab result variable by a normal distribution, then initialize all patients' healthy data $X_{init}$ accordingly. 

\noindent\textbf{Stage II: Disease Assignment.}
Randomly set up diseases and the DAG causal graph, then build up causality function from diseases to lab results, i.e., diseases influence labs value. Generate all patients' diseased data $X_{dise}$ accordingly.


\noindent\textbf{Stage III: Treatment Assignment.}
Randomly set up multiple medicine lines for each disease, and build up the causality function from medicines to the responded disease, i.e., treatments affect disease severity. Update all patients' treated data $X_{trea}$ accordingly, and in each iteration, record the difference of after and before updating to be the value of $\Delta$. 



\vspace{-2mm}
\begin{figure}[h!]
\includegraphics[width=1.0\columnwidth]{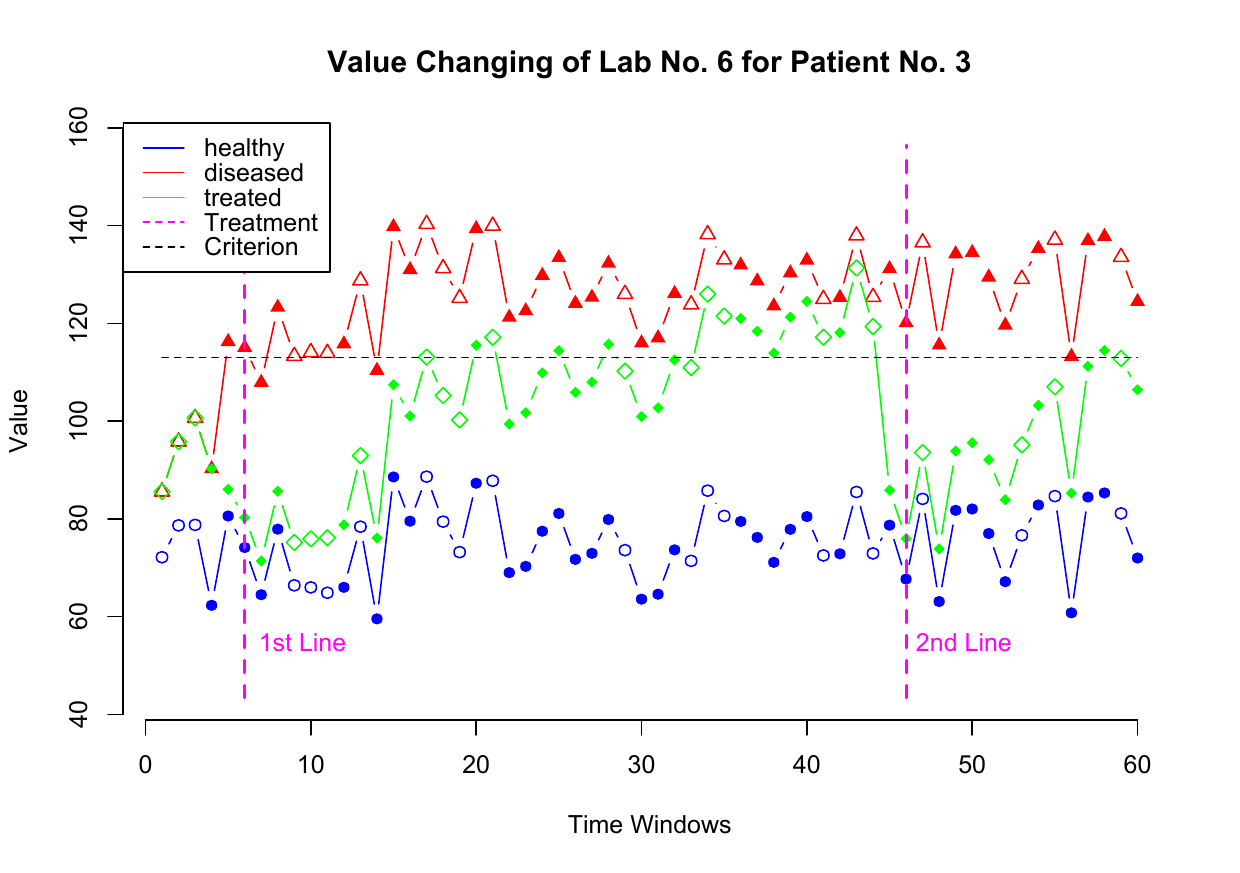}
\vspace{-.3in}
\caption{Example of synthetic data. The filled and unfilled markers indicate the random hospital visiting and absence, simulated by Poisson distribution. }
\label{fig:syn_plot}
\vspace{-2mm}
\end{figure}

As the example displayed in Fig.\ref{fig:syn_plot}, for patient ``No.3", the three views of his values on ``Lab No.6" are colored by blue, red, and green, respectively, and all of them keep changing along the timeline. This patient developed a sequence of diseases (not shown in the figure) in a causal order, whose influence on ``Lab No.6" is reflected as the value changing of the red curve. To cure him, the treatments, i.e., medicines, are assigned based on the observed values of the red curve (by pre-determined criterion), and the value changing of the green curve reflects the treatment effects. Each disease has a sequence of functional medicines, ordered by increasing levels, and the $i$th level is named as ``$i$th Line''.

We published the introduced causal data generator for free downloading\footnote
{https://github.com/kflijia/Causal-EHR-Generator}, implemented on the R platform.
As a reminder, the generator is on chronic disease simulation, which develops as a group of comorbidities, such as diabetes, hypertension, and heart disease. Notably, the disease severity $d\in D$ naturally progresses non-decreasing along the timeline; And all medications in $M$ are kept ongoing, i.e., no stop once start taken. It does not apply to generating pandemic diseases.

\begin{table*}[t!]
\centering
\caption{ Disease Diagnosis Prediction Performance of $4$ Different Outcomes on Real EHR Dataset}
\label{tab:real_class_res}
\begin{tabular}{|c|c|c|c|c|c|c|c|c|c|c|c|}
\hline
\multicolumn{12}{|c|}{AUC Performance} \\ \hline
\multicolumn{2}{|c|}{Models} & \multicolumn{2}{c|}{GLM} & \multicolumn{2}{c|}{GLMER} & \multicolumn{3}{c|}{LSTM} & \multicolumn{3}{c|}{GRU} \\ \hline
\multicolumn{2}{|c|}{Methods} & \multicolumn{1}{l|}{No $\Delta$} & \multicolumn{1}{l|}{$\Delta$ Augment} & \multicolumn{1}{l|}{No $\Delta$} & \multicolumn{1}{l|}{$\Delta$ Augment} & \multicolumn{1}{l|}{No $\Delta$} & \multicolumn{1}{l|}{$\Delta$ Augment} & \multicolumn{1}{l|}{$\Delta$ PreTrain} & \multicolumn{1}{l|}{No $\Delta$} & \multicolumn{1}{l|}{$\Delta$ Augment} & \multicolumn{1}{l|}{$\Delta$ PreTrain} \\ \hline
\multirow{4}{*}{T2DM} & \multicolumn{1}{l|}{Step=1} & 0.7783 & 0.7806**\hspace{1.5mm} & 0.7778 & 0.7803**\hspace{1.5mm} & 0.7829 & 0.7005\hspace{4mm} & 0.8067*** & 0.7846 & 0.7019\hspace{4mm} & \textbf{0.8069}*** \\ \cline{2-12} 
 & \multicolumn{1}{l|}{Step=2} & 0.8000 & 0.8041*** & 0.7488 & 0.7529*** & 0.8325 & 0.7387\hspace{4mm} & 0.8448*** & 0.8437 & 0.7459\hspace{4mm} & \textbf{0.8588}*** \\ \cline{2-12} 
 & \multicolumn{1}{l|}{Step=3} & 0.6867 & 0.6942*** & 0.6483 & 0.6533*** & 0.6888 & 0.6782\hspace{4mm} & \textbf{0.7083}*** & 0.6778 & 0.6665\hspace{4mm} & 0.7054*** \\ \cline{2-12} 
 & \multicolumn{1}{l|}{Step=4} & 0.6722 & 0.6756*** & 0.6712 & 0.6758*** & 0.6559 & 0.6845*** & 0.6861*** & 0.6564 & 0.6722*\hspace{2.9mm} & \textbf{0.6893}*** \\ \hline
\multirow{4}{*}{CAD} & Step=1 & 0.7059 & \textbf{0.7202}*** & 0.6998 & 0.7156*** & 0.6868 & 0.6577\hspace{4mm} & 0.7061*** & 0.6874 & 0.6607\hspace{4mm} & 0.7104*** \\ \cline{2-12} 
 & Step=2 & 0.7127 & 0.7233*** & 0.6805 & 0.6887*** & 0.7122 & 0.6707\hspace{4mm} & 0.7213*** & 0.7169 & 0.6692\hspace{4mm} & \textbf{0.7292}*** \\ \cline{2-12} 
 & Step=3 & 0.6376 & 0.6418*** & 0.6259 & 0.6280*** & 0.6121 & 0.6310*** & 0.6337*** & 0.6049 & 0.6320*** & \textbf{0.6372}*** \\ \cline{2-12} 
 & Step=4 & 0.6475 & 0.6527*** & 0.6487 & \textbf{0.6538}*** & 0.6079 & 0.6418*** & 0.6454*** & 0.6059 & 0.6494*** & 0.6486*** \\ \hline
\multirow{4}{*}{CKD} & Step=1 & 0.8154 & 0.8159\hspace{4.5mm} & 0.8154 & \textbf{0.8159}\hspace{4.5mm} & 0.7685 & 0.6470\hspace{4mm} & 0.7877*** & 0.7684 & 0.6469\hspace{4mm} & 0.7942*** \\ \cline{2-12} 
 & Step=2 & 0.8182 & \textbf{0.8200}*** & 0.7792 & 0.7814*** & 0.7990 & 0.6632\hspace{4mm} & 0.8094*** & 0.7899 & 0.6603\hspace{4mm} & 0.8147*** \\ \cline{2-12} 
 & Step=3 & 0.7816 & \textbf{0.7829}*\hspace{2.9mm} & 0.7456 & 0.7469*\hspace{3mm} & 0.7352 & 0.6539\hspace{4mm} & 0.7596*** & 0.7327 & 0.6661\hspace{4mm} & 0.7660*** \\ \cline{2-12} 
 & Step=4 & 0.7687 & 0.7698*\hspace{3mm} & 0.7720 & \textbf{0.7731}*\hspace{3mm} & 0.6802 & 0.6814\hspace{4mm} & 0.7299*** & 0.6775 & 0.6816\hspace{4mm} & 0.7332*** \\ \hline
\multirow{4}{*}{Stroke} & Step=1 & 0.6409 & 0.6395\hspace{4.4mm} & 0.6409 & 0.6394\hspace{4mm} & 0.6165 & 0.6143\hspace{4mm} & \textbf{0.6754}*** & 0.6115 & 0.6242\hspace{4mm} & 0.6675*** \\ \cline{2-12} 
 & Step=2 & 0.6393 & 0.6388\hspace{4.4mm} & 0.6454 & 0.6482\hspace{4mm} & 0.6053 & 0.6118\hspace{4mm} & \textbf{0.6642}*** & 0.6011 & 0.6113*\hspace{2.9mm} & 0.6634*** \\ \cline{2-12} 
 & Step=3 & 0.6343 & 0.6443*** & 0.6519 & 0.6412\hspace{4mm} & 0.6024 & 0.6233*\hspace{3mm} & \textbf{0.6614}*** & 0.6059 & 0.6104\hspace{4mm} & 0.6528*** \\ \cline{2-12} 
 & Step=4 & 0.6491 & 0.6565**\hspace{1.5mm} & 0.6453 & 0.6550**\hspace{1.5mm} & 0.6123 & 0.6532*** & \textbf{0.7013}*** & 0.6044 & 0.6588*** & 0.6882*** \\ \hline 
\end{tabular}
\end{table*}

\begin{table*}[]
\centering
\caption{ Labs Value Forecast Performance of $3$ Different Outcomes on Real EHR Dataset}
\label{tab:real_regr_res}
\resizebox{2\columnwidth}{!}{%
\begin{tabular}{|c|c|c|c|c|c|c|c|c|c|c|c|}
\hline
\multicolumn{12}{|c|}{MSE Performance} \\ \hline
\multicolumn{2}{|c|}{Models} & \multicolumn{2}{c|}{LM} & \multicolumn{2}{c|}{LMER} & \multicolumn{3}{c|}{LSTM} & \multicolumn{3}{c|}{GRU} \\ \hline
\multicolumn{2}{|c|}{Methods} & \multicolumn{1}{l|}{No $\Delta$} & \multicolumn{1}{l|}{$\Delta$ Augment} & \multicolumn{1}{l|}{No $\Delta$} & \multicolumn{1}{l|}{$\Delta$ Augment} & \multicolumn{1}{l|}{No $\Delta$} & \multicolumn{1}{l|}{$\Delta$ Augment} & \multicolumn{1}{l|}{$\Delta$ PreTrain} & \multicolumn{1}{l|}{No $\Delta$} & \multicolumn{1}{l|}{$\Delta$ Augment} & \multicolumn{1}{l|}{$\Delta$ PreTrain} \\ \hline
\multirow{4}{*}{LDL} & \multicolumn{1}{l|}{Step=1} & 969.53 & 908.32*** & 991.06 & 932.49*** & 1006.04 & 917.85*** & \textbf{749.58}*** & 983.1 & 903.26*** & 750.38*** \\ \cline{2-12} 
 & \multicolumn{1}{l|}{Step=2} & 964.51 & 885.84*** & 970.12 & 893.03*** & 987.93 & 941.46**\hspace{1.5mm} & 714.05*** & 986.62 & 902.53**\hspace{1.5mm} & \textbf{546.27}*\hspace{3mm} \\ \cline{2-12} 
 & \multicolumn{1}{l|}{Step=3} & 946.47 & 792.52*** & 955.84 & 810.55*** & 1104.67 & 903.83*** & 729.19*** & 996.01 & 1012.31\hspace{4mm} & \textbf{516.05}*** \\ \cline{2-12} 
 & \multicolumn{1}{l|}{Step=4} & 827.67 & 681.55*** & 827.66 & \textbf{681.41}*** & 866.8 & 836.14*** & 689.75*** & 1092.29 & 818.17*\hspace{3mm} & 863.58*** \\ \hline
\multirow{4}{*}{SBP} & Step=1 & 466.74 & 457.74*** & 468.02 & 460.37*** & 512.51 & 479.64*** & 433.21*** & 512.98 & 484.75*** & \textbf{421.41}*** \\ \cline{2-12} 
 & Step=2 & 432.93 & 418.55*** & 430.61 & 418.09*** & 488.42 & 458.56*** & \textbf{396.06}*** & 488.7 & 473.07*\hspace{3mm} & 418.13*** \\ \cline{2-12} 
 & Step=3 & 385.72 & 367.77*** & 385.2 & 368.41*** & 444.67 & 419.25**\hspace{1.5mm} & \textbf{343.38}*** & 459.5 & 464.11\hspace{4.5mm} & 375.98*** \\ \cline{2-12} 
 & Step=4 & 358.91 & \textbf{337.74}**\hspace{1.5mm} & 358.91 & 337.74**\hspace{1.5mm} & 479.9 & 460.52**\hspace{1.5mm} & 361.04**\hspace{1.5mm} & 507.64 & 445.96*** & 361.31*** \\ \hline
\multirow{4}{*}{TG} & Step=1 & 5262.75 & 4754.05*** & 5732.52 & 5425.32*** & 5465.9 & 4732.7*** & 4230.93*** & 5428.54 & 4661.68*** & \textbf{4090.07}*** \\ \cline{2-12} 
 & Step=2 & 6105.08 & 5339.08*** & 6556.5 & 5807.74*** & 6349.99 & 5365.36*** & 4812.35*** & 6506.7 & 5300.33*** & \textbf{4523.19}*** \\ \cline{2-12} 
 & Step=3 & 4571.69 & 3971.84*** & 4538.45 & 4026.53*** & 4864.27 & 4497.26*** & \textbf{3487.98}**\hspace{1.5mm} & 4745.23 & 4373.88*** & 4478.10*** \\ \cline{2-12} 
 & Step=4 & 4102.99 & 3342.96*** & 4105.74 & 3347.37*** & 4761.85 & 4217.16*** & \textbf{2223.00}**\hspace{1.5mm} & 5654.48 & 5128.49*\hspace{3mm} & 3593.77*\hspace{3mm} \\ \hline
\end{tabular}
}
\end{table*}

\vspace{1mm}
\subsubsection{Synthetic Setting}
Mainly, in our experiments, we set up the observational timeline as $40$ years, i.e., $40$ time steps in total, and a varying number of synthetic patients are sampled by demand (from $500$ to $5000$). 

There are $20$ lab results, whose values are always under observation along the timeline, and 10 diseases (denoted as ``$d_1,\ldots,d_{10}$'') that each disease can possibly influence $1\sim 4$ labs value. These $10$ diseases form a group of comorbidities, and with a pre-determined causal graph, $G$ exists at most $20$ possible disease-developing paths (i.e., disease trajectories).

We set up $10$ classes of medicines corresponding to the $10$ diseases. Each class contains $1\sim 3$ medicine levels, ordered with increasing impacts on lab value, including both positive typical and adverse side effects.

\vspace{1mm}
\subsubsection{Prediction Tasks}
The experiments on synthetic data aim to exhibit the predictive performance changing 
by varying hyper-parameter settings. The parameters include:
1) Time Step (i.e., the number of observed preceding years); 2) Training data sample size (i.e., the number of synthetic patients).

From these results, we can confirm the conditions in which the proposed $\Delta$ can be most effective, i.e., the most helpful to enhance the predictive performance. Additionally, we can observe whether involving $\Delta$ would influence the computing efficiency, i.e., running time of the training process.

For well concentrating, unlike in the actual data experiments, we only perform the binary classification tasks: To predict the 5-years-risk of the synthetic disease diagnosis, i.e., whether the patient will develop the disease in future $5$ years.

We adopt AUC (Area under the ROC Curve) as the performance evaluation. For each model, we compare the two AUCs with and without $\Delta$ input. Additionally, the training time is also recorded and the comparison is reached at the end.

\vspace{1mm}
\subsubsection{Compared Models}
\begin{itemize}
    \item GLM, i.e. Logistic Linear Regression Model.
    \item RNN-GRU in Unsupervised Pre-training Architecture.
    \item RNN-LSTM in Unsupervised Pre-training Architecture.
\end{itemize}

Here we dropped the Linear Mixed-Effects Model and the $\Delta$-Augmentation RNNs, because they did not show out-performance compared to their brother methods.

We expect: 1) RNN-GRU and RNN-LSTM can outperform the linear model GLM in most of the tasks, especially with large sample size; 2) adding $\Delta$ input can significantly enhance predictive performance for both linear model and RNN model; 3) The training time is expected to be longer with $\Delta$ added, but not dramatically.

\section{Experimental Results}
\label{sec:results}



\vspace{-1mm}
For real EHR data, all experiments are conducted for $20$ times, with $20$ randomly splitting of training and test. For synthetic EHR data, we independently generate $5$ different data,
and perform all experiments for $4$ times on each one, with $4$ randomly splitting.

All shown results are on averaged values over these multiple independent runs, and the corresponding conference intervals (CI) are also displayed, as the shadowed areas in figures.

The splitting rate is $0.7:0.3$ for training and test.

\vspace{-1mm}
\subsection{Results on Real Diabetes EHR Data}
\vspace{-0.5mm}

\begin{table}[]
\begin{adjustwidth}{-2mm}{}
\caption{P-values of t-test on AUC vectors pair}
\label{tab:ttest_pvalue}
\resizebox{1.05\columnwidth}{!}{
\renewcommand{\arraystretch}{1.2}
\begin{tabular}{|p{7mm}|c|c|c|c|c|c|}
\hline
\multicolumn{2}{|c|}{$10^{-3} *$} & \multicolumn{5}{c|}{Sample Size} \\ \hline
Models & Time Steps & 250 & 500 & 1500 & 2500 & 5000 \\ \hline
\multirow{3}{*}{GLM} & Step=5 & 3.2903 & 1.1429 & 0.6845 & 0.7187 & 0.7336 \\ \cline{2-7} 
 & Step=10 & 0.2065 & 0.9230 & 0.0682 & 0.3762 & 0.0130 \\ \cline{2-7} 
 & Step=15 & 1.5441 & 2.6313 & 1.0823 & 2.4657 & 0.8256 \\ \hline
\multirow{3}{*}{GRU} & Step=5 & 0.0000 & 0.0000 & 0.0015 & 0.0000 & 0.1455 \\ \cline{2-7} 
 & Step=10 & 0.0021 & 0.0046 & 0.0347 & 0.2035 & 0.0486 \\ \cline{2-7} 
 & Step=15 & 0.3203 & 0.4357 & 0.1109 & 0.6493 & 4.0279 \\ \hline
\multirow{3}{*}{LSTM} & Step=5 & 0.0005 & 0.0000 & 0.0008 & 0.0003 & 0.0001 \\ \cline{2-7} 
 & Step=10 & 0.0229 & 0.4461 & 1.1835 & 1.2984 & 8.7623 \\ \cline{2-7} 
 & Step=15 & 0.0651 & 2.6749 & 9.9602 & 12.3447 & 39.8459 \\ \hline
\end{tabular}
}
\end{adjustwidth}
\vspace{-2mm}
\end{table}

Table \ref{tab:real_class_res} and \ref{tab:real_regr_res} display the performances of binary diagnosis predictions (in AUC) and the continuous labs value forecasts (in MSE) respectively. 
Each row in the Table represents an individual task with a specified outcome and Time Step. The columns indicate applied methods, and the bold text in each row indicates the best performer. 
The star symbols stand for significance level of the improvement with $\Delta$'s participation, compared to the no $\Delta$ version. The improvement is verified by one-side paired t-tests, 
and significance levels are defined as p-value intervals: $***$ for $(-\infty, 0.001]$, $**$ for $(0.001, 0.01]$, and $*$ for $(0.01,0.05]$.

In Table \ref{tab:real_class_res} and \ref{tab:real_regr_res}, the RNN methods with $\Delta$ pre-training architecture, including both RNN-GRU and RNN-LSTM (but especially LSTM), outperformed the others mostly. Specifically, they won 10 times out of 16 binary classification tasks and 10 times out of 12 continuous labs value forecasting tasks. On the other hand, RNN methods without $\Delta$ input never outperformed the pre-training ones.

Except for RNNs, $\Delta$ can also benefit linear models. Most of the time, with $\Delta$ augmented, linear models performed better but not as significantly as RNNs.

\subsection{Results On Synthetic EHR Data}
\vspace{-1mm}

\begin{figure}[h!]
\hspace{-7.5mm}
\includegraphics[scale=0.68]{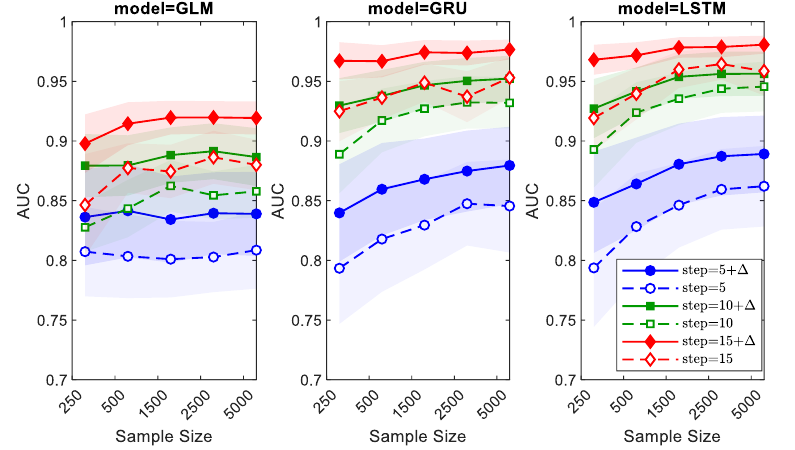}
\caption{ Comparison of the predictive performances with increasing numbers of samples for different Time Steps.}
\vspace{-3mm}
\label{fig:res_auc}
\end{figure}

Fig\ref{fig:res_auc} shows AUC performances of the three compared models. The predicted outcomes are five randomly selected synthetic diseases, whose performances are equally weighted.
%
For each model, with and without $\Delta$ performances are in the same color but distinguished by solid and dashed line style. 
With varying Time Step=$5,10,15$, the sample size increases from $250$ to $5000$ gradually. Shadows in figure represent CI.
As we expected, the performance with $\Delta$ is always better than the one without $\Delta$ input. And both RNN-GRU and RNN-LSTM consistently perform better than GLM when the sample size is large enough (in our case $>=500$).

The hyper-parameters (Time Step and sample size) obviously influence the performance. With larger Time Step, all models tend to perform better, due to more historical information observed; With increasing sample size, RNN models present an upward trend on performance, but level off when sample size has been large enough. On the other hand, GLM model slightly performs the upward trend but much weaker than RNNs, especially with small time Step $=5$.

\begin{figure}[h!]
\hspace{-7.5mm}
\includegraphics[scale=0.68]{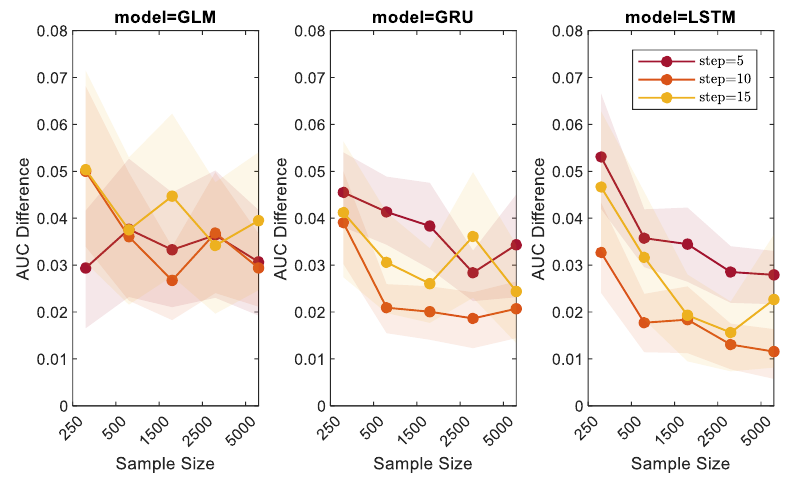}
\caption{ Performance gains by adding $\Delta$ for three methods: GLM, GRU, and LSTM.}
\label{fig:res_auc_diff}
\vspace{-4mm}
\end{figure}

Figure \ref{fig:res_auc_diff} shows the effect of adding $\Delta$. It includes the AUC gain of involving $\Delta$, i.e. the difference of with-$\Delta$-AUC minus without-$\Delta$-AUC, and keeps consistent layout with Figure \ref{fig:res_auc}. 
With shrinking sample size (from 5000 to 250), RNN-LSTM clearly presents the trend that $\Delta$ becomes more helpful on improving predictions; RNN-GRU trends the same but less obviously; And this pattern is almost vanished on GLM model. On the other hand, increasing Time Step from 5 to 15 did not result in any monotonic tendency. 

For more convincing conclusions, we performed paired t-test on each pair of AUC vectors. 
Each vector consists of the performances of all  synthetic disease prediction tasks. The vector here has a length $80$. The p-values are listed in Table\ref{tab:ttest_pvalue}, and all turn out to be significant. 
Particularly, RNN-LSTM shows a clear pattern that when the sample size is comparatively small, $\Delta$ can be more certainly to be helpful, which is consistent with the conclusion from Figure \ref{fig:res_auc_diff}.

Among the three models, RNN-LSTM acts the most stably to present $\Delta$'s effect changing. Because LSTM has the most complex architecture, and the largest number of parameters, it is more able to capture enough information from data than the other two. In contrast, the simplest model GLM failed to fully utilize the new information brought by $\Delta$.

On the side of training efficiency, adding $\Delta$ does not dramatically influence it. 
For GLM, augmentation of $\Delta$ increased training time by around $80\%$, e.g. with sample size=$5000$ and Time Step=$15$, training without $\Delta$ runs $10.8$ secs by average and increased to $18.82$ secs with $\Delta$.
RNN models spend hundreds of secs on the training process,
and after pre-training using $\Delta$, their training time reduced by around $1/3$ because of prior knowledge. However, the entire time cost is almost doubled compared to the version without pre-training.

\section{Conclusion}\label{sec:conclusion}
In this paper, we proposed a novel method to quantitatively estimate the domain-defined treatment effects from EHR data as a generalized feature vector, which is simply utilizable for any properly selected machine learning methodology.

We demonstrated the theoretical analysis about the principle of our method and proved that the proposed estimator comes with new causal information, which can hardly be automatically discovered by regular RNN models. 
Furthermore, with comprehensive experiments we have displayed that our method can effectively help Deep Learning models to leverage clinical knowledge that is originally hidden in EHR data; and consequentially improved their predictive performance, especially when the sample size is comparatively small.

We believe that our works have taken the Deep Learning applications a step closer to a form of interpretable Artificial Intelligence integrated Healthcare.






\begin{appendices}
\section{Synthetic EHR Data Generation Algorithms}

The algorithms below are referred to in Section 5.2.1 . 
And all used notations are listed in Table\ref{tab:notation}.
Synthetic EHR data generation comprises three stages: I) Initialization, II) Disease Assignment, III) Treatment Assignment.

\begin{table}[h!]
\caption{Notations in Causal Data Generation Algorithm}
\label{tab:notation}
  \begin{center}
    \label{tab:table1}
    \resizebox{\columnwidth}{!}{%
    \begin{tabular}{|l|l|}
    \hline
       $t= 1\ldots T$ & $T$ time windows\\
       $P=\{p\}_{N_p} \mid p\in \mathcal{P}$ & The set of $N_p$ patients in space $\mathcal{P}$\\
       $L=\{l\}_{N_l} \mid l\in \mathcal{L}$ & The set of $N_l$ lab results in space $\mathcal{L}$\\
       $D=\{d\}_{N_d} \mid d\in \mathcal{D}$ & The set of $N_d$ diseases in space $\mathcal{D}$\\
       $M=\{m\}_{N_m} \mid m\in \mathcal{M}$ & The set of $N_m$ medicine in space $\mathcal{M}$\\
       $e(d_1,d_2) \mid d_1,d_2\in D$ & A causal relation from $d_1$ to $d_2$. \\
       $G(D,\{e\})$ & The causal graph of diseases $D$.\\
       $V(G)=D$ & Function of getting vertices of $G$. \\
       $\mathcal{L}=\mathbf{F}_d(\mathcal{D})$ & The causation mapping from $\mathcal{D}$ to $\mathcal{L}$.\\
       $\mathcal{D}=\mathbf{F}_m(\mathcal{M})$ & The causation mapping from $\mathcal{M}$ to $\mathcal{D}$.\\
       $X$ and $\Delta$ & Data matrix and $\Delta$ matrix.\\
      \hline
    \end{tabular} %
    }
  \end{center}
\end{table}

\noindent\textbf{Stage I: Initialization.}
Randomly set up the healthy values for each lab result variable by a normal distribution, then initialize all patients' healthy data $X_{init}$ accordingly. 

\vspace{-3mm}
\begin{algorithm}
\footnotesize
\SetAlgoLined
\KwResult{ $(N_p \times T \times N_l)$ data matrix $X_{init}$ }
\For{$p= (1\ldots N_p)$ }{  
    \tcp{For each patient $p$}
    \For{$l= (1\ldots N_l)$}{    
        \tcp{For each lab result $l$}
        Random $(\mu,\sigma)$ \;
        Random $(x^1,\ldots,x^T)$ as $x^t\sim\mathcal{N}(\mu,\sigma), t=1\ldots T$\;
        $X_{init} [p,:,l]= (x^1,\ldots,x^T)$\;
    }
}
 \caption{(Stage I) Initialization}
\end{algorithm}

\vspace{-3mm}
\noindent\textbf{Stage II: Disease Assignment.}
Randomly set up diseases and the DAG causal graph of them, then build up causality function from diseases to lab results, i.e. diseases influence labs value. Generate all patients' diseased data $X_{dise}$ accordingly.

\vspace{-3mm}
\begin{algorithm}
\footnotesize
\SetAlgoLined
\KwResult{$(N_p \times T \times N_l)$ data matrix $X_{dise}$ }
Initialize $X_{dise}=X_{init}$\;
Random graph $G=(D,\{e\});$ 
\tcp{Diseases causal graph}
Random $\mathcal{L}=\mathbf{F}_d(\mathcal{D});$
\tcp{Causation from $\mathcal{D}$ to $\mathcal{L}$}
\For{$p= (1\ldots N_p)$}{
    \tcp{For each patient $p$}
    Random path $g$ from $G$ \;
    \For{$d\in V(g)$}{
        \tcp{For each $d$ in path}
        Update $L$ by $L=\mathbf{F}_d(d);$
        \tcp{Update lab results}
        \For{$l\in L$}{
            Update $X_{dise} [p,:,l];$
            \tcp{Update time series}
        }
    }
}
 \caption{(Stage II) Disease Assignment}
\end{algorithm}

\vspace{-3mm}
\noindent\textbf{Stage III: Treatment Assignment.}
Randomly set up multiple medicine lines for each disease, and build up the causality function from medicines to the responded disease, i.e. treatments affect disease severity. Update all patients' treated data $X_{trea}$ accordingly, and in each iteration, record the difference of after and before updating to be the value of $\Delta$.

\begin{algorithm}
\footnotesize
\SetAlgoLined
\KwResult{$(N_p \times T \times N_l)$ data matrix $X_{trea}$ and $\Delta$ }
Initialize $X_{trea}=X_{dise}$\;
Initialize $\Delta$ as zeros\;
Random $\mathcal{D}=\mathbf{F}_m(\mathcal{M});$
\tcp{Causation from $\mathcal{M}$ to $\mathcal{D}$}
\For{$p= (1\ldots N_p)$}{
    \tcp{For each patient $p$}
    Random medicine set $M$ from $\mathcal{M}$ \;
    \For{$m\in M$}{
        Update $D$ by $D=\mathbf{F}_m(m);$
        \tcp{Update diseases}
        \For{$d\in D$}{
             Update $L$ by $L=\mathbf{F}_d(d);$
            \tcp{Update labs}
            \For{$l\in L$}{
                \tcp{Update time series}
                Record $x_0 = X_{trea} [p,:,l]$\;
                Update $X_{trea} [p,:,l]$ \;
                Update $\Delta [p,:,l]=X_{trea} [p,:,l] - x_0$ \;
            }
        }
    }
}
 \caption{(Stage III) Treatment Assignment}
\end{algorithm}

\end{appendices}

\end{document}